\documentclass[sigconf]{acmart}
\usepackage{url}

\usepackage{multirow}
\usepackage[marginal]{footmisc}
\usepackage{balance}

\AtBeginDocument{%
  \providecommand\BibTeX{{%
    \normalfont B\kern-0.5em{\scshape i\kern-0.25em b}\kern-0.8em\TeX}}}

\copyrightyear{2021} \acmYear{2021} \setcopyright{acmcopyright}\acmConference[MM '21]{Proceedings of the 29th ACMInternational Conference on Multimedia}{October 20--24, 2021}{Virtual Event, China}\acmBooktitle{Proceedings of the 29th ACM International Conference on Multimedia(MM '21), October 20--24, 2021, Virtual Event, China}\acmPrice{15.00}\acmDOI{10.1145/3474085.3475574}\acmISBN{978-1-4503-8651-7/21/10}

\begin{document}
\fancyhead{}

\title{Learning Multi-Granular Spatio-Temporal Graph Network for Skeleton-based Action Recognition}






\author{Tailin Chen$^{1\ast\dagger}$, Desen Zhou$^{2\ast}$, Jian Wang$^{2}$, Shidong Wang$^{1}$, Yu Guan$^{1}$, Xuming He$^{3\ddagger}$, Errui Ding$^{2}$}




\affiliation{
\institution{$^{1}$Open Lab, Newcastle University, Newcastle upon Tyne, UK}
\institution{$^{2}$Department of Computer Vision Technology (VIS), Baidu Inc., China}
\institution{$^{3}$ShanghaiTech University, Shanghai, China}
\institution{\{t.chen14, shidong.wang, yu.guan\}@newcastle.ac.uk}
\institution{\{zhoudesen, wangjian33, dingerrui\}@baidu.com, hexm@shanghaitech.edu.cn}
\city{}
\country{}
}

\thanks{$\ast$ Equal contribution. $\ddagger$ Corresponding Author. $\dagger$ Work done when Tailin Chen was a research intern at Baidu VIS and visiting student at ShanghaiTech University.}







\begin{abstract}
The task of skeleton-based action recognition remains a core challenge in human-centred scene understanding due to the multiple granularities and large variation in human motion. 
Existing approaches typically employ a single neural representation for different motion patterns, which has difficulty in capturing fine-grained action classes given limited training data. To address the aforementioned problems, we propose a novel multi-granular spatio-temporal graph network for skeleton-based action classification that jointly models the coarse- and fine-grained skeleton motion patterns. 
To this end, we develop a dual-head graph network consisting of two interleaved branches, which enables us to extract features at two spatio-temporal resolutions in an effective and efficient manner. Moreover, our network utilises a cross-head communication strategy to mutually enhance the representations of both heads. 
We conducted extensive experiments on three large-scale datasets, namely NTU RGB+D 60, NTU RGB+D 120, and Kinetics-Skeleton, and achieves the state-of-the-art performance on all the benchmarks, which validates the effectiveness of our method\footnotemark[1].
\end{abstract}





\maketitle

\renewcommand{\thefootnote}{\fnsymbol{footnote}}
\footnotetext{$^{1}$Code available at \textcolor{magenta}{\url{https://github.com/tailin1009/DualHead-Network}}}






\section{Introduction}

Action recognition is a fundamental task in human-centred scene understanding and has achieved much progress in computer vision and multimedia. Recently, skeleton-based action recognition has attracted increasing attention to the community due to the advent of inexpensive motion sensors \cite{cao2018openpose,ntu_60} and effective human pose estimation algorithms \cite{st-gcn,2s-agcn,ms-g3d}. The skeleton data typically are more compact and robust to environment conditions than its video counterpart, and accurate action recognition from skeletons can greatly benefit a wide range of applications, such as human-computer interactions, healthcare assistance and physical education. 

\begin{figure}[tbp]
    \includegraphics[width=0.5\textwidth]{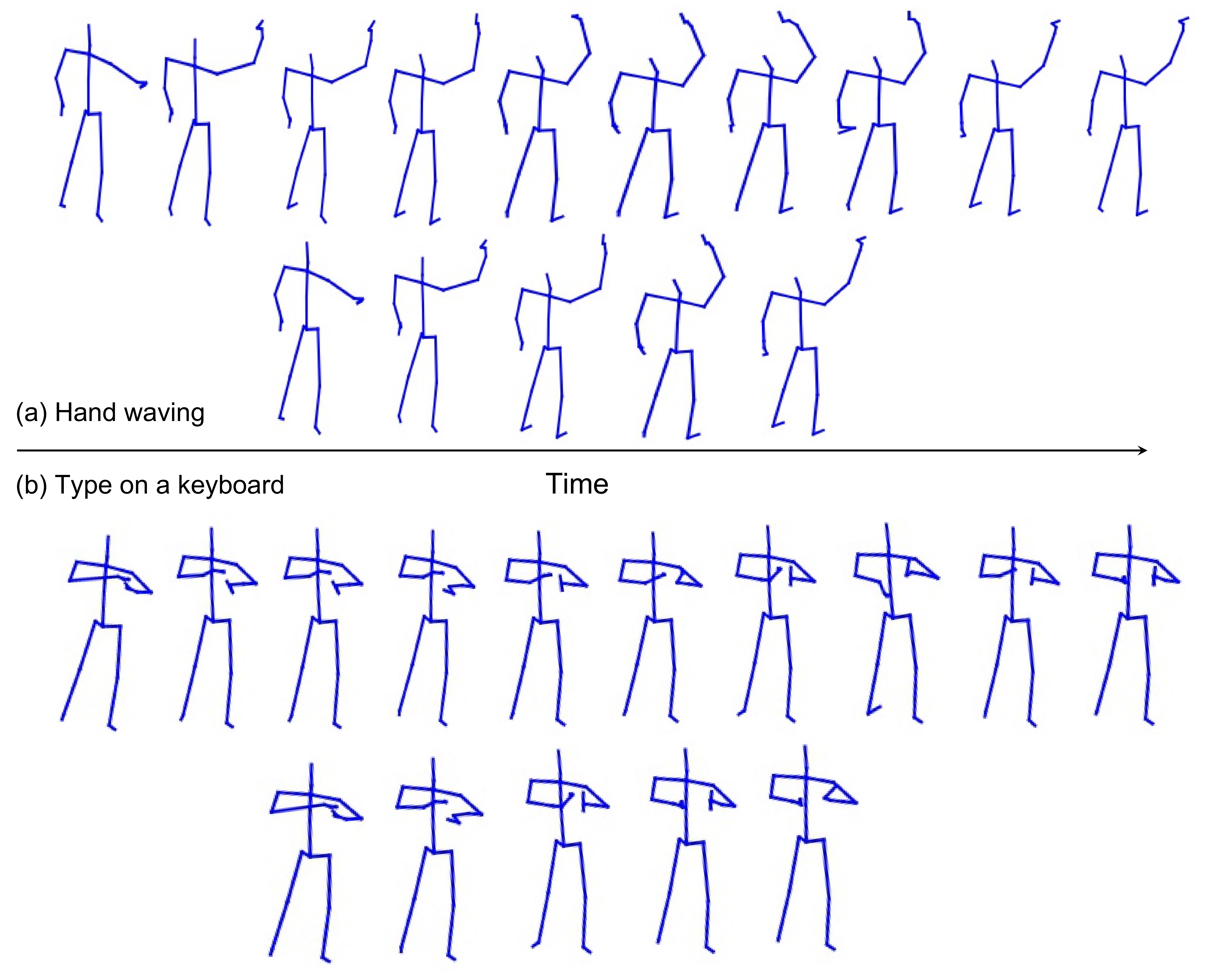}
    \vspace{-0.3cm}
	\caption{\small Some actions can be recognised via coarse-grained temporal motion patterns, such as `hand waving' (top), while recognising other actions requires not only coarse-grained motion but subtle temporal movements, such as `type on a keyboard' (bottom).}
	\vspace{-0.2cm}
	\label{fig:advertisment}
	\Description{}
\end{figure}

Different from RGB videos, skeleton data contain only the 2D or 3D coordinates of human joints, which makes skeleton-based action recognition particularly challenging due to the lack of image-based contextual information.
In order to capture discriminative spatial structure and temporal motion patterns, existing methods \cite{ms-aagcn,st-gcn,ms-g3d} usually rely on a shared spatio-temporal representation for the skeleton data at the original frame rate of input sequences. While such a strategy may have the capacity to capture action patterns of multiple scales, they often suffer from inaccurate predictions on fine-grained action classes due to high model complexity and limited training data. To tackle this problem, we argue that a more effective solution is to explicitly model the motion patterns of skeleton sequences at multiple temporal granularity.  
For example, actions such as `hand waving' or `stand up' can be distinguished based on coarse-grained motion patterns, while recognizing actions like `type on a keyboard' or `writing' requires understanding not only coarse-grained motions but also subtle temporal movements of pivotal joints, as shown in Figure.\ref{fig:advertisment}. To the best of our knowledge, such multiple granularity of temporal motion information remains implicit in the recent deep graph network-based approaches, which is less effective in practice.


In this paper, we propose a dual-head graph neural network framework for skeleton-based action recognition in order to capture both coarse-grained and fine-grained motion patterns. Our key idea is to utilise two branches of interleaved graph networks to extract features at two different temporal resolutions. The branch with lower temporal resolution captures motion patterns at a coarse level, while the branch with higher temporal resolution is able to encode more subtle temporal movements. Such coarse-level and fine-level feature extraction are processed in parallel and finally the outputs of both branches are fused to perform dual-granular action classification.

Specifically, we first exact a base feature representation of the input skeleton sequence by feeding it into a backbone Graph Convolution Network (GCN). 
Then we perform two different operations to the resulting feature maps: the first operation subsamples the feature maps at the temporal dimension with a fixed downsampling rate, which removes the detailed motion information and hence produces a coarse-grained representation; in the second operation, we keep the original temporal resolution and utilise an embedding function to generate a fine-grained representation. 

Subsequently, we develop two types of GCN modules to process the resulting coarse- and fine-grained representations, which are referred as \textit{fine head} and \textit{coarse head}. Each head consists of two sequential GCN blocks, which extract features within respective granularity. In particular, our coarse head captures the correlations between joints at a lower temporal resolution, hence infers actions in a more holistic manner. 
To facilitate such coarse-level inference, we estimate a temporal attention from the fine-grained features in the fine head, indicating the importance of each frame. The temporal attention is used to re-weight the features at the coarse head. 
The intuition behind such cross head attention is as follows: here we pass the fine-grained motion contexts encoded in the attention to the coarse head in order to remedy the lack of fine level information in the coarse head. 
Similarly, we utilise the coarse-grained features to estimate a spatial attention indicating the importance of joints. The spatial attention highlights the pivotal joint nodes in the fine head. 
Our fine GCN blocks are able to focus on the subtle temporal movements of the pivotal joints and hence extract the fine-grained information effectively. 
Finally, each head predicts an action score, and the final prediction is given by fusion of two scores.

We validate our method by extensive experiments on three public datasets: NTU RGBD+D 60\cite{ntu_60}, NTU RGB+D 120\cite{ntu_120}, Kinetics-Skeleton\cite{kinetics}. The results show that our method outperforms existing works in all the benchmarks, demonstrating the effectiveness of the proposed coarse-and-fine dual head structure. To summarize, our contributions are three-folds:





1) We propose a dual-head spatio-temporal graph network that can effectively learn robust human motion representations at both coarse- and fine-granularity for skeleton-based action recognition.

2) We design a cross head attention mechanism to mutually enhance the spatio-temporal features at both levels of granularity, which enables the model to focus on key motion information.

3) Our dual-head graph neural network achieves new state-of-the-art on three public benchmarks.

\begin{figure*}[!t]
	\centering
	\includegraphics[width=0.95\textwidth]{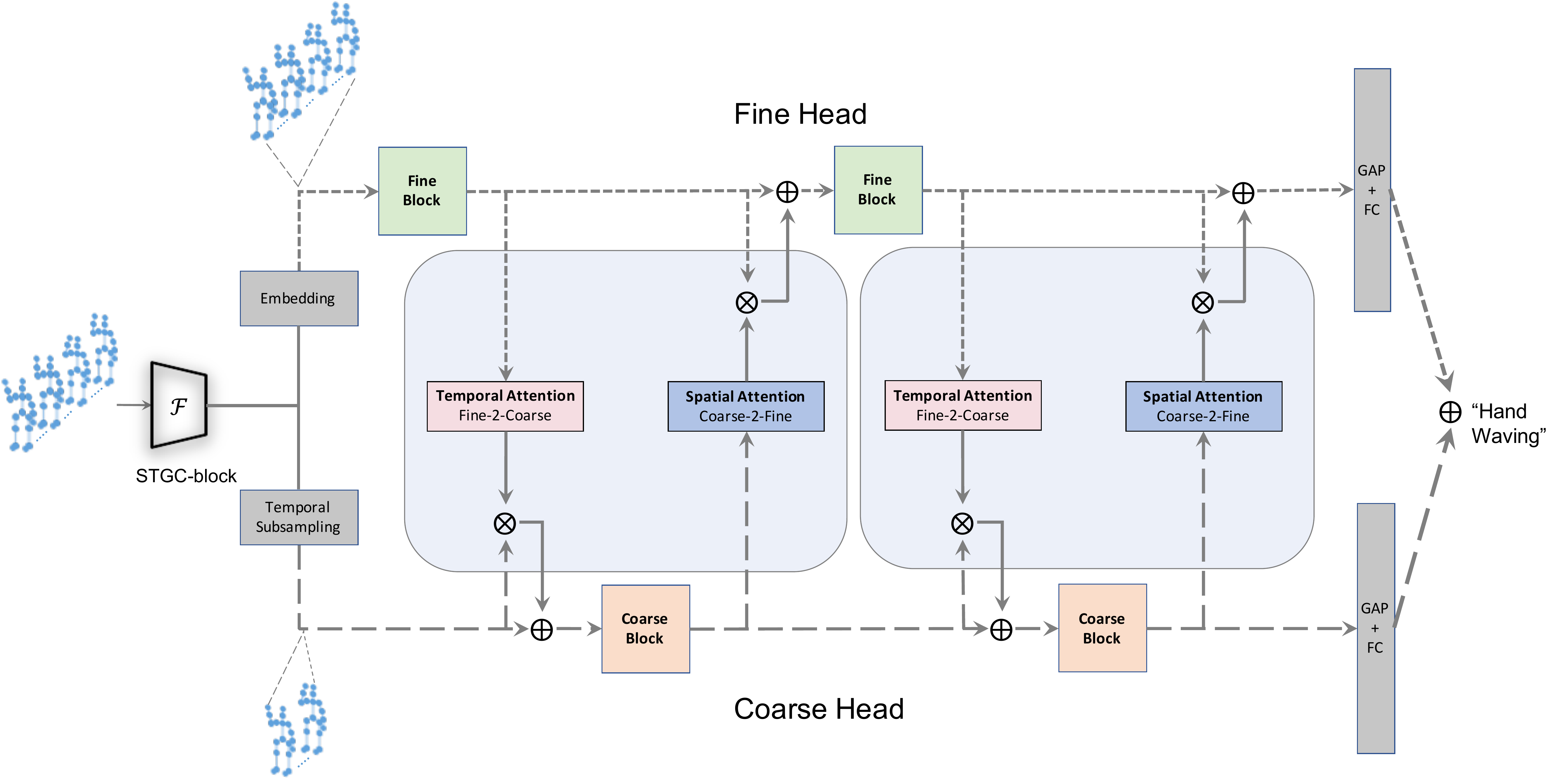}
	\vspace{-0.3cm}
	\caption{\small Overview of our proposed framework. We first utilise a STGC block to generate backbone features of skeleton data, and then use a coarse head to capture coarse-grained motion patterns, and a fine head to encode fine-grained subtle temporal movements. Cross-head spatial and temporal attentions are exploited to mutually enhance feature representations. Finally each head generates a probabilistic prediction of actions and the ultimate estimation is given by fusion of both predictions.}
	\vspace{-0.4cm}
	\label{fig:overview}
	\Description{}
\end{figure*}



\section{Related Work}
 
\subsection{Skeleton-based Action Recognition}

Early approaches on skeleton-based action recognition typically adopt hand-crafted features to capture the human body motion \cite{vemulapalli2014human,hu2015jointly,hussein2013human}. However, they mainly rely on exploiting the relative 3D rotations and translations between joints, and hence suffer from complicated feature design and sub-optimal performance. Recently, deep learning methods have achieved significant progress in skeleton-based action recognition, which can be categorized into three groups according to their network architectures: i.e., Recurrent Neural Networks (RNNs), Convolutional Neural Networks (CNNs) and Graph Convolutional Networks (GCNs).



The RNN-based methods usually first extract frame-level skeletal features and then model sequential dependencies with RNN models  \cite{du2015hierarchical,ntu_60,liu2016spatio, li2017adaptive, zhang2017view}. For example, \cite{li2017adaptive} constructed an adaptive tree-structured RNN while \cite{zhang2017view} designed a view adaptation scheme for modeling actions. LSTM networks have also been employed to mine global informative joints \cite{liu2017global}, to extract co-occurrence features of skeleton sequences \cite{zhu2016co}, or to learn stacked temporal dynamics~\cite{si2018skeleton}. 
The CNN-based methods typically convert the skeleton sequences into a pseudo-image and employ a CNN to classify the resulting image into action categories. In particular, \cite{li2018co} designed a co-occurrence feature learning framework and \cite{kim2017interpretable} used a  one-dimensional residual CNN to identify skeleton sequences based on directly-concatenated joint coordinates. \cite{liu2017enhanced} proposed 10 types of spatio-temporal images for skeleton encoding, which are enhanced by visual  and  motion features. 
However, both the RNNs and CNNs have difficulty in capturing the skeleton topology which are naturally of a graph structure.

To better capture human body motion, recent works utilise GCNs \cite{st-gcn,2s-agcn} for spatial and temporal modeling of actions. A milestone of the GCN-based method is ST-GCN\cite{st-gcn}, which defines a sparse connected spatial-temporal graph that both considers natural human body structure and temporal motion dependencies in space-time domain. Since then, a large body of works adopt the GCNs for skeleton-based action recognition:  2s-AGCN\cite{2s-agcn} proposed an adaptive attention module to learn non-local dependencies in spatial dimension. \cite{zhang2020context} explored contextual information between joints. \cite{ms-aagcn} further introduced spatial and temporal attentions in the GCN. \cite{shift-gcn} developed a shift convolution on the graph-structured data. \cite{ms-g3d} proposed a multi-scale aggregation method which can effectively aggregate the spatial and temporal features. \cite{dynamic_gcn} designed a context-enriched module for better graph connection. \cite{ResGCN} utilised multiple data modality in a single model. \cite{MST-GCN-AAAI2021} proposed multi-scale spatial temporal graph for long range modeling. Nevertheless, those existing methods rely on a single shared representation for multi-granular motion, which is difficult to learn from limited training data. By contrast, our method explicitly design a two-branch graph network to explicitly capture coarse- and fine-grained motion patterns and hence is more effective on modeling action classes.

\subsection{Video-based Action Recognition}
In video-based action recognition, several methods have explored temporal modeling at multiple pathways or temporal resolutions. SlowFast Networks \cite{feichtenhofer2019slowfast} ``factorize'' spatial and temporal inference in two different pathways. Temporal Pyramid Network\cite{yang2020temporal} explored various visual tempos by fusing features at different stages with multiple temporal resolutions. Recent Coarse-Fine networks\cite{kahatapitiya2021coarse} propose a re-sampling module to improve long-range dependencies. However, multiple-level motion granularity are not sufficiently explored in skeleton-based action recognition. Due to the lack of image level context, we need to carefully design the networks for each level of granularity. In addition, we propose a cross-head attention strategy to enhance each head with another, resulting in an effective dual-granular reasoning.

\section{Methods}

\subsection{Overview}

Our goal is to jointly capture the multi-granular spatio-temporal dependencies in skeleton data and learn discriminative representations for action recognition. To this end, we develop a novel dual-head network that explicitly captures motion patterns at two different spatio-temporal granular levels. Each head of our network adopts a different temporal resolution and hence focuses on extracting a specific type of motion features. In particular, a fine head maintains the original temporal resolutions as the input so that it can model fine-grained local motion patterns, while a coarse head uses a lower temporal resolution via temporal subsampling so that it can focus more on coarse-level temporal contexts. Moreover, we further introduce a cross-head attention module to ensure that the extracted information from different heads can be communicated in a mutually reinforcing way. The dual-head network generates its final prediction by fusing the output scores of both heads. 

The details of the proposed method are organised as follows. Firstly, we introduce the GCNs (Sec.~\ref{sec:data}) and backbone module (Sec.~\ref{sec:backbone}) for skeletal feature extraction. Secondly, we depict the dual-head module (Sec.~\ref{sec:dual-head}), the cross-communication attention module (Sec.~\ref{sec:cross}) and the fusion module (Sec.~\ref{sec:fuse}). Finally, we describe the details of the multi-modality ensemble strategy (Sec.~\ref{sec:multi}).

\subsection{GCNs on Skeleton Data}\label{sec:data}
The proposed framework adopts graph convolutional networks to effectively capture the dependencies between dynamic skeleton joints. Below we introduce three basic network modules used in our method, including MS-GCN, MS-TCN and MS-G3D.

Formally, given a skeleton of $N$ joints, we define a skeleton graph as $\mathcal{G}=(\mathcal{V}, \mathcal{E})$, where $\mathcal{V} = \{v_1,...,v_{N}\}$ is the set of $N$ joints and $\mathcal{E}$ is the collection of edges. The graph connectivity can be represented by the adjacency matrix $\mathbf{A} \in \mathbb{R}^{N \times N}$, where its element value takes 1 or 0 indicating whether the positions of $v_{i}$ and $v_{j}$ are adjacent. 
Given a skeleton sequence, we first compute a set of features $\mathcal{X}=\{{x}_{t,n}|1\leq t \leq T,1\leq n \leq N; n, t \in \mathbb{Z}\}$ that can be represented as a feature tensor $\textbf{X} \in \mathbb{R}^{C \times T \times N}$, where $x_{t,n}=\textbf{X}_{t,n}$ denotes the \textit{C} dimensional features of the node $ v_{n} $ at time \textit{t}. 

\paragraph{\noindent\textbf{MS-GCN}} The spatial graph convolutional blocks(GCN) aims to capture the spatial correlations between human joints within each frame $\mathbf{X}_t$. We adopt a multi-scale GCN(MS-GCN) to jointly capture multi-scale spatial patterns in one operator:

    	\begin{equation}\label{eq1}
    	\textbf{Y}_t^\mathcal{S} = \sigma \left( \sum_{k=0}^{K} \widetilde{\textbf{A}}_{k} \textbf{X}_t\textbf{W}_{k} \right),
    	\end{equation}
where $ \textbf{X}_t \in \mathbb{R}^{N \times C}$ and $ \textbf{Y}_t^\mathcal{S} \in \mathbb{R}^{N \times C_{\rm{out}}} $ denote the input and output features respectively. Here $\textbf{W}_k \in \mathbb{R}^{C \times C_{\rm{out}}}$ are the graph convolution weights, and \textit{K} indicates the number of scales of the graph to be aggregated. $\sigma(\cdot)$ is the activation function. $\widetilde{\textbf{A}}_k$ are the normalized adjacency matrices as in \cite{kipf2016semi,li2019actional} and can be obtained by: $\widetilde{\textbf{A}}_k=\hat{\textbf{D}}_k^{-\frac{1}{2}}\hat{\textbf{A}}_k\hat{\textbf{D}}_k^{-\frac{1}{2}}$, where $\hat{\textbf{A}}_k=\textbf{A}_k+\textbf{I}$ is the adjacency matrix including the nodes of the self-loop graph ( \textbf{I} is the identity matrix) and $\hat{\textbf{D}}_k$ is the degree matrix of $\textbf{A}_k$. We denote the output of entire sequence as $\textbf{Y}^\mathcal{S}=[\textbf{Y}_1^\mathcal{S},\cdots,\textbf{Y}_T^\mathcal{S}]$.

\paragraph{\noindent\textbf{MS-TCN}} The temporal convolution(TCN) is formulated as a classical convolution operation on each joint node across frames. We adopt multiple TCNs with different dilation rates to capture temporal patterns more effectively. A $N$-scale MS-TCN with kernel size of $K_t \times 1$ can be expressed as, 
	   \begin{equation}\label{eq2}
    	\textbf{Y}_\mathcal{T} = \sum_{i}^{N} Conv2D[K_t \times 1; D^i](\textbf{X}).
    	\end{equation}
where $D^i$ denotes the dilatation rate of $i^{th}$ convolution.

\paragraph{\noindent\textbf{MS-G3D}} To jointly capture spatio-temporal patterns, a unified graph operation(G3D) on space and time dimension is used. We also adopt a multi-scale G3D in our model. Please refer to \cite{ms-g3d} for more detailed descriptions.

\subsection{Backbone Module}\label{sec:backbone}
We first describe the backbone module of our network, which computes the base features of the input skeleton sequence. In this work, we adopt the multi-scale spatial-temporal graph convolution block (STGC-block) \cite{ms-g3d}, which has proven effective in representing long-range spatial and temporal context of the skeletal data. 

Specifically, given an input sequence $\{z_{t,n}\in \mathbb{R}^d| 1\leq t \leq T, 1\leq n \leq N; t,n \in \mathbb{Z}\}$, where $d\in \{2,3\}$ indicates the dimension of joint locations, the output of the backbone module can be defined as: 
\begin{equation}
    \mathcal{X}_{back} = \{x^{(back)}_{t,n}\in \mathbb{R}^{\mathcal{C}_{back}} | 1\leq t\leq T, 1\leq n\leq N; t,n\in\mathbb{Z}\},
\end{equation}
where $\mathcal{C}_{back}$ is the channel dimension of the output feature, and $x^{(back)}_{t,n}$ indicates the representation of a specific joint $n$ at frame $t$.


\begin{figure}[!tbp]
	\includegraphics[width=0.45\textwidth]{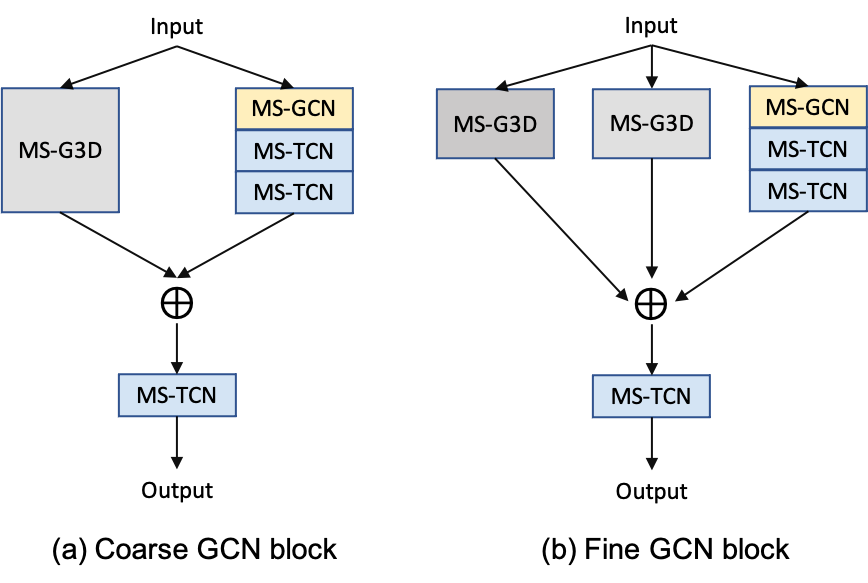}
	\vspace{-0.4cm}
	\caption{\small Basic blocks in coarse head(left) and fine head(right). We simplify the GCN blocks in two directions. In coarse block, we reduce the G3D component with the largest window size to reduce temporal modeling; in fine block, we reduce the convolution kernels though channel dimensions to 1/2 of coarse block.}
	\label{fig:GCNBlocks}
	\vspace{-0.4cm}
\end{figure}

\subsection{Dual-head Module}\label{sec:dual-head}

To capture the motion patterns with inherently variable temporal resolutions, we develop a multi-granular spatio-temporal graph network. In contrast to prior work relying on shared representations, we adopt a dual-head network to simultaneously extract motion patterns from coarse and fine levels. Given the backbone features $\mathcal{X}_{back}$, below we introduce the detailed structure of our coarse head and fine head. 

\paragraph{\noindent\textbf{Coarse Head}} Our coarse head extracts features at a low temporal resolution, aiming to capture coarse grained motion contexts. In the coarse head, a subsampling layer is first adopted to downsample the feature map at the temporal dimension. Concretely, given the backbone features $\mathcal{X}_{back}$ with $T$ frames, 
we uniformly sample $T/\alpha$ nodes in the temporal dimension:
\begin{equation}
    \mathcal{X}_{coar} = \mathcal{F}_{subs}(\mathcal{X}_{back}),
\end{equation}
where $\mathcal{F}_{subs}$ and $\alpha\in \mathbb{Z}$ denote the subsampling function and subsampling rate respectively. $\mathcal{X}_{coar}= \{x^{(coar)}_{t,n}\in \mathbb{R}^{\mathcal{C}_{coar}}|1\leq t\leq T/\alpha, 1\leq n\leq N; t,n\in\mathbb{Z}\}$ represents the initial feature maps of coarse head.

Subsequently, we introduce a coarse GCN block, denoted by $\mathcal{G}_{coar}$, which consists of two parallel paths formed by a MS-G3D and stacking of multiple MS-GCN and MS-TCN respectively, followed by a MS-TCN fusion block. The detailed structures of coarse GCN block is shown in Figure \ref{fig:GCNBlocks} (a). The coarse GCN block is used to compute the final coarse feature representation as follows:
\begin{equation}
    \widetilde{\mathcal{X}_{coar}} = \mathcal{G}_{coar}(\mathcal{X}_{coar})
\end{equation}
where $\widetilde{\mathcal{X}_{coar}}$ is the output feature set. 
 
\paragraph{\noindent\textbf{Fine Head}}  Our fine head extracts features at a high temporal resolution and encode more fine-grained motion contexts. In the fine head, an embedding function $\mathcal{F}_{embed}$ (i.e., $1 \times 1$ convolution layer) will be applied at the beginning to reduce the feature dimensions. In this way, the output features of the backbone module $\mathcal{X}_{back}$ will be projected into the new feature space $\mathcal{X}_{fine}$ :
\begin{equation}
    \mathcal{X}_{fine} = \mathcal{F}_{embed}(\mathcal{X}_{back}),
\end{equation}
where $\mathcal{X}_{fine} = \{x^{(fine)}_{t,n}\in \mathbb{R}^{\mathcal{C}_{fine}}|1\leq t\leq T, 1\leq n\leq N; t,n\in\mathbb{Z}\}$.
Then we introduce a fine GCN block, denoted as $\mathcal{G}_{fine}$, to extract fine-grained temporal features as below,
\begin{equation}
    \widetilde{\mathcal{X}_{fine}} = \mathcal{G}_{fine}(\mathcal{X}_{fine}).
\end{equation}
The fine GCN block consists of three parallel branches formed by two MS-G3D and stacking of multiple MS-GCN and MS-TCN respectively, followed by a MS-TCN fusion block.The detailed structures of fine GCN block is shown in Figure \ref{fig:GCNBlocks} (b).


\paragraph{\noindent\textbf{Block Simplification}} The proposed dual-head network, a novel structure based on the divide-and-conquer strategy, can effectively extract motion patterns of different levels of granularity in the temporal domain. 
Note that, due to the downsampling operation, the temporal receptive field has already been expanded. As a result, GCN blocks in such structure naturally can be simplified, such as using smaller temporal convolution kernels. 
Specifically, in the coarse head, we remove G3D component with the largest window size in the STGC-block to reduce temporal modeling. In the fine head, as the original temporal resolution data already maintains rich temporal contexts, we then reduce the channel dimensions for efficient modelling. The detailed structures of coarse GCN block and fine GCN block are shown in Figure \ref{fig:GCNBlocks}.

\subsection{Cross-head Attention Module}\label{sec:cross}
To better fuse the representations at different temporal resolutions, we introduce a cross-head attention module to enable communication between two head branches. Such communication can mutually enhance the representations encoded by both head branches.  



\begin{figure}[!tbp]
	\includegraphics[width=0.4\textwidth]{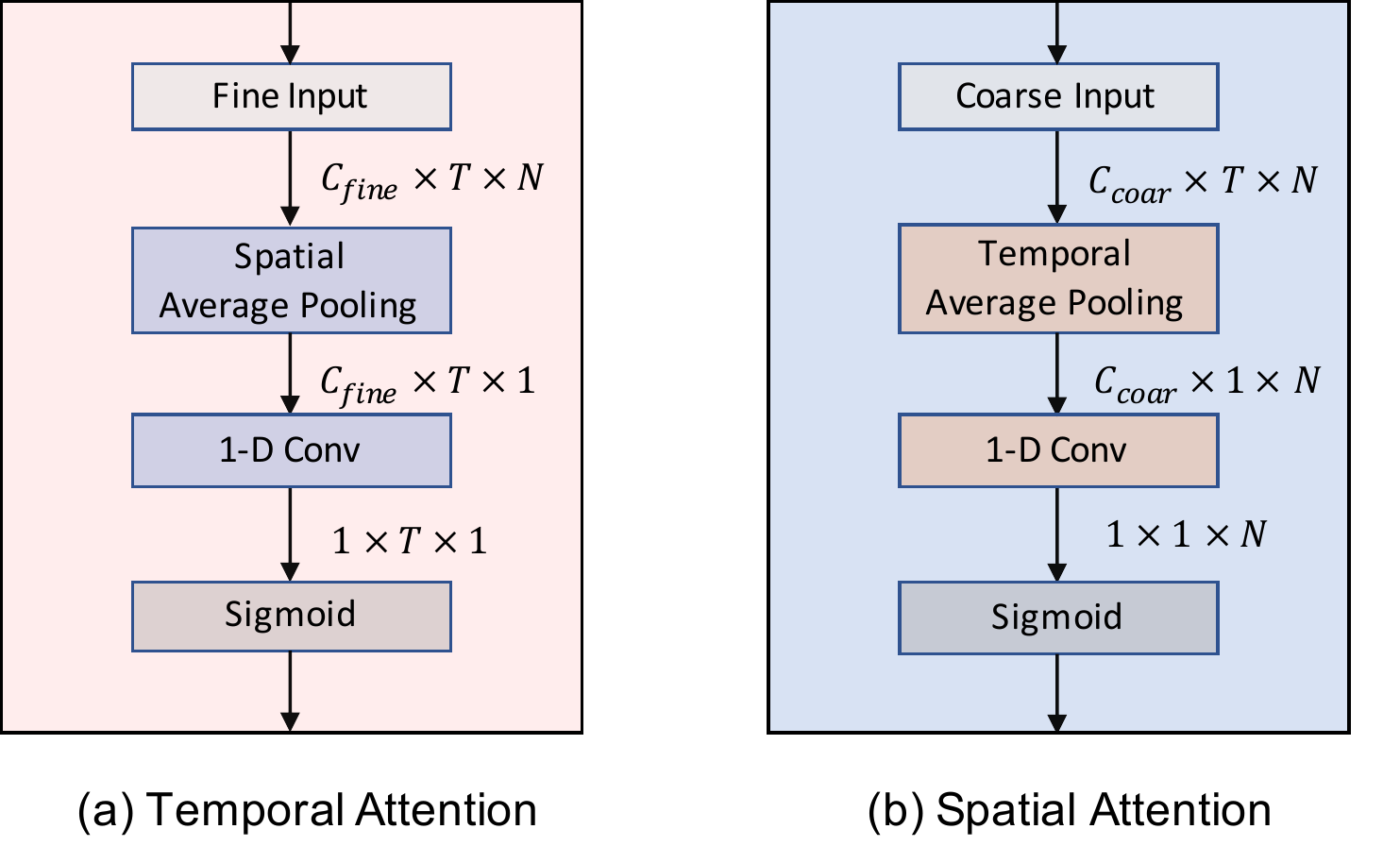}
	\vspace{-0.3cm}
	\caption{\small An illustration of our spatial and temporal attention block.}
\label{fig:Spatial_temporal_attention_module}
\vspace{-0.5cm}
\end{figure}

\paragraph{\noindent\textbf{Communication Strategy}}
The detailed workflow of the proposed communication strategy can be found in Figure \ref{fig:overview}. Since the unsampled frames intrinsically contain more elaborate motion patterns than the downsampled frames, the granular temporal information will be initially transmitted from the fine head to the coarse head. Taking the first temporal attention block as an example, the output of the Fine block will go though an attention block similar to the SE network \cite{hu2018squeeze}. 
The generated attention serves as an interactive message, and then the correlation with the coarse temporal feature is obtained through element-wise matrix multiplication. The correlated feature will be finally fused with the coarse temporal feature and fed into the coarse block for the upcoming propagation. The spatial attention holds a similar structure as the temporal attention, but the order of input features is different. The detailed calculation of these two attention will be introduced below and the structure is shown in Figure \ref{fig:Spatial_temporal_attention_module}.

\paragraph{\noindent\textbf{Temporal Attention}}
The temporal attention block takes the output of the fine block as the input and can extract the temporal patterns with high similarity from the distant frames to the greatest extent because it retains the complete frame rate of the input sequence. The features learned from this can fully reflect the importance of the frame level and lead the coarse-level reasoning in an efficient way of communication. Formally, it can be denoted as: 
\begin{equation}
    \theta_{te} = \sigma(\mathbf{W_{te}}(AvgPool_{sp}(\mathcal{X}_{fine}))),
\end{equation}
where $\mathcal{X}_{fine}$ is the feature map in fine head, $AvgPool_{sp}$ denotes the average pool in spatial dimension, $\mathbf{W_{te}}$ indicates a 1D convolution with a large kernel size to capture large temporal receptive field. $\sigma$ indicates sigmoid activation function, $\theta_{te}\in\mathbb{R}^{1\times T\times 1}$ is the estimated temporal attention. The attention is used to re-weight coarse features $\mathcal{X}_{coar}$ in a residual manner: 
\begin{equation}
    \hat{\mathcal{X}}_{coar} = \theta_{te}\cdot\mathcal{X}_{coar} + \mathcal{X}_{coar},
    \label{update_te_atten}
\end{equation}
where $\cdot$ indicates the element-wise matrix multiplication with shape alignment.

\paragraph{\noindent\textbf{Spatial Attention}} 


Our fine head aims to extract motion patterns from subtle temporal movements. To promote such fine-grained representation learning, we then utilise the spatial attention to highlight important joints across frames. Compared with fine-head itself, our coarse head extracts features in lower temporal resolution and can easily learn high-level abstractions in a holistic view. We therefore take the advantage of such holistic representation from coarse head to estimate the spatial attention. 
Formally, 
\begin{equation}
    \theta_{sp} = \sigma(\mathbf{W_{sp}}(AvgPool_{te}(\mathcal{X}_{coar}))),
\end{equation}
where $AvgPool_{te}$ indicates the average pooling at temporal dimension, $\mathbf{W_{sp}}$ indicates a 1D convolution layer, $\theta_{sp}\in\mathbb{R}^{1\times1\times N}$ is the joint level attention, which is used to re-weight fine features in a residual manner:
\begin{equation}
    \hat{\mathcal{X}}_{fine} = \theta_{sp} \cdot \mathcal{X}_{fine} + \mathcal{X}_{fine},
    \label{update_sp_atten}
\end{equation}

As shown in Figure \ref{fig:overview}, our temporal attention and spatial attention are alternately predicted to enhance temporal and spatial features of both heads.

\subsection{Fusion Module}\label{sec:fuse}
The proposed network has two heads, responsible for the different granularities of temporal reasoning. We utilise the score level fusion to combine the information of both head and facilitate the final prediction. 
For simplicity, we denote the outputs of coarse head and fine head as $\widetilde{\mathcal{X}_{coar}}$ and $\widetilde{\mathcal{X}_{fine}}$, and then each head is attached with a global average pooling (GAP) layer, a fully connected layer combined with a SoftMax function to predict a classification score $s_{coar}$ and $s_{fine}$:

\begin{equation}
    s_{coar} = SoftMax(\mathbf{W_{coar}}(AvgPool(\widetilde{\mathcal{X}_{coar}}))),
\end{equation}
\begin{equation}
    s_{fine} = SoftMax(\mathbf{W_{fine}}(AvgPool(\widetilde{\mathcal{X}_{fine}}))).
\end{equation}

where $s_{coar}, s_{fine}\in\mathbb{R}^{a}$ indicate the estimated probability of $a$ classes in the dataset, $AvgPool$ is a GAP operation conducted on the spatial and temporal dimensions, $\mathbf{W_{coar}}$ and $\mathbf{W_{fine}}$ are fully connected layeres. 
Then, the final prediction can be achieved by fusion of these two scores:
\begin{equation}
    s = \mu \cdot s_{coar} + (1-\mu) \cdot s_{fine},
\end{equation}
where $\mu $ is a hyperparameter used to examine the importance between two heads. During our implementation, we empirically set $\mu=0.5$. More configurations can be explored in future work. During traning, we also supervise $s_{coar}$ and $s_{fine}$ with two cross entropy losses and sums them with the same weight $\mu$.

\subsection{Multi-modality Ensemble}\label{sec:multi}
Following prior works\cite{2s-agcn,ms-aagcn,shift-gcn,MST-GCN-AAAI2021}, we generate four modalities for each skeleton sequence, they are the joint, bone, joint motion and bone motion. Specifically, the joint modality is derived from the raw position of each joints. The bone modality is produced by the offset between two adjacent joints in a predefined human structure. The joint motion modality and bone motion modality are generated by the offsets of the joint data and bone data in two adjacent frames. We recommend that readers refer to \cite{2s-agcn,ms-aagcn} for more details about multi-stream strategy. Our final model is a 4-stream network that sums the softmax scores of each modality to make predictions.

\section{Experiments}

\subsection{Datasets}

\textbf{NTU RGB+D 60.} NTU RGB+D 60 \cite{ntu_60} is a large-scale indoor-captured action recognition dataset, containing 56,880 skeleton sequences of 60 action classes captured from 40 distinct subjects and 3 different camera perspectives. Each skeleton sequence contains the 3D spatial coordinates of 25 joints captured by the Microsoft Kinect v2 cameras. Two evaluation benchmarks are provided, (1) Cross-Subject (X-sub): half of the 40 subjects are used for training, and the rest are used for testing. (2) Cross-View (X-view): the samples captured by cameras 2 and 3 are selected for training and the remaining samples are used for testing.\vspace{0.05cm}

\noindent\textbf{NTU RGB+D 120.}
NTU RGB+D 120 \cite{ntu_120} is an extension of the NTU RGB+D 60\cite{ntu_60} in terms of the number of performers and action categories and currently is the largest 3D skeleton-based action recognition dataset containing 114,480 action samples from 120 action classes. 
Two evaluation benchmarks are provided, (1) Cross-Subject (X-sub) that splits 106 subjects into training and test sets, where each set contains 53 subjects; (2) Cross-setup (X-set) that splits the collected samples by the setup IDs (i.e., even setup IDs for training and odds setup IDs for testing).\vspace{0.05cm}

\noindent \textbf{Kinetics-Skeleton.}
Kinetics dataset \cite{kinetics} contains approximately 300,000 video clips in 400 classes collected from the Internet. The skeleton information is not provided by the original dataset but estimated by the publicly available Open-Pose toolbox \cite{cao2018openpose} . The captured skeleton information contains 18 body joints, as well as their 2D coordinates and confidence score. There are 240,436 samples for training and 19,794 samples for testing. 


\begin{table}[t!]
		\centering
		\scalebox{1}{
			\begin{tabular}{lcccc}
				\toprule
				\multirow{2}{*}{Methods} & \multirow{2}{*}{Publisher} &\multicolumn{2}{c}{NTU RGB+D 60} \\ 
				\cline{3-4} &  & X-sub     &   X-view   \\  
				\midrule
				GCA-LSTM \cite{liu2017global}           &CVPR17    &  74.4 & 82.8     \\
				VA-LSTM \cite{zhang2017view}           &ICCV17    &  79.4 & 87.6     \\
				TCN \cite{kim2017interpretable}         &CVPRW17  &  74.3 & 83.1     \\
				Clips+CNN+MTLN \cite{ke2017new}         &CVPR17    &  79.6 & 84.8     \\
				\midrule
				ST-GCN \cite{st-gcn}                    &AAAI18   &  81.5 & 88.3     \\
				SR-TSL \cite{si2018skeleton}            &ECCV18   &  84.8 & 92.4     \\

				STGR-GCN \cite{li2019spatio}            &AAAI19    &  86.9 & 92.3       \\
				AS-GCN \cite{li2019actional}            &CVPR19    &  86.8 & 94.2     \\
				2s-AGCN \cite{2s-agcn}                  &CVPR19    &  88.5 & 95.1     \\
				DGNN \cite{dgcn}             &CVPR19    &  89.9 & 96.1      \\
				GR-GCN \cite{gao2019optimized}          &ACMMM19   &  87.5 & 94.3     \\
				SGN \cite{zhang2020semantics}           &CVPR20    &  89.0 & 94.5     \\
				MS-AAGCN \cite{ms-aagcn}                &TIP20     &  90.0 & 96.2       \\
				NAS-GCN \cite{peng2020learning}         &AAAI20    &  89.4 & 95.7     \\
				4s Decouple-GCN \cite{decouple_gcn_cheng2020eccv}   &ECCV2020    &  90.8 & \textbf{96.6}     \\

                4s Shift-GCN \cite{shift-gcn}           &CVPR20    &  89.7 & \textbf{96.6}     \\
                STIGCN \cite{huang2020spatio}        &ACMMM20   &  90.1 & 96.1     \\
                ResGCN \cite{ResGCN}                    &ACMMM20   &  90.9 & 96.0     \\
                4s Dynamic-GCN \cite{dynamic_gcn}          &ACMMM20   &  91.5 & 96.0     \\
                
				2s MS-G3D \cite{ms-g3d}                    &CVPR20    &  91.5 & 96.2 \\  
				4s MST-GCN \cite{MST-GCN-AAAI2021}              &AAAI21   & 91.5 & \textbf{96.6}     \\
				\midrule
				Js DualHead-Net (Ours)        &    -    & 90.3        & 96.1   		\\
				Bs DualHead-Net (Ours)        &    -   & 90.7        & 95.5  		\\
				2s DualHead-Net (Ours)        &    -    & \textbf{91.7 }       & 96.5  		\\
				4s DualHead-Net (Ours)        &    -   & \textbf{92.0 }      & \textbf{96.6}   		\\
				\bottomrule
		\end{tabular}}
		\caption{\small Comparison of the Top-1 accuracy (\%) with the state-of-the-art methods on the NTU RGB+D 60 dataset.}
		\vspace{-0.5cm}
		\label{ntu60}
\end{table}

\subsection{Implementation Details}
All experiments are conducted using PyTorch. The cross-entropy loss is used as the loss function, and Stochastic Gradient Descent (SGD) with Nesterov Momentum (0.9) is used for optimization. The downsample ratio of coarse head is set to $\beta=2$. The fusion weight $\mu$ of the two heads is set to $0.5$ and the weight $\lambda$ of the loss function is set to $1$.

The preprocessing of the NTU-RGB+D 60\&120 dataset is in line with previous work \cite{ms-g3d}. 
During training, the batch size is set to 64 and the weight decay is set to 0.0005. The initial learning rate is set to 0.1, and then divided by 10 in the $40^{th}$ epoch and $60^{th}$ epoch. The training process ends in the $80^{th}$ epoch. 
The experimental setup of the Kinetics-Skeleton dataset is consistent with previous works \cite{st-gcn,2s-agcn}.
We set the batch size to 128 and weight decay to 0.0001. The learning rate is set to 0.1 and is divided by 10 in the $45^{th}$ epoch and the $55^{th}$ epoch. The model is trained for a total of 80 epochs. 



\subsection{Comparisons with the State-of-the-Art Methods}
We compare the proposed method (DualHead-Net) with the state-of-the-art methods on three public benchmarks. Following the previous works \cite{2s-agcn,ms-aagcn,shift-gcn,MST-GCN-AAAI2021}, we generate four modalities data (joint, bone, joint motion and bone motion) 
and report the results of joint stream (Js), bone stream (Bs), joint-bone two-stream fusion (2s), and four-stream fusion (4s). Experimental results are shown in Table\ref{ntu60}, Table \ref{ntu120} and Table \ref{kinetics} respectively. On three large-scale datasets, our method outperforms existing methods under all evaluation settings. 

Specifically, as shown in Table \ref{ntu60}, our method achieves state-of-the-art performance \textbf{91.7} on \textbf{NTU RGB+D 60} X-sub setting with only joint-bone two-stream fusion. The final four-stream model further improves the performance to \textbf{92.0}.
For \textbf{NTU RGB+D 120} (Table \ref{ntu120}), it is worth noting that on X-sub setting, our single-stream (Bs) model is competitive to two-stream baseline MS-G3D\cite{ms-g3d}, which demonstrates the effectiveness of our proposed dual-head graph network design.


Furthermore, for the largest \textbf{Kinetics-Skeleton}, as shown in Table \ref{kinetics}, our four-stream model outperforms prior work\cite{MST-GCN-AAAI2021} by \textbf{0.3} in terms of the top-1 accuracy.

	\begin{table}[t!]
		\centering
		\scalebox{1}{
			\begin{tabular}{lcccc}
				\toprule
				\multirow{2}{*}{Methods} & \multirow{2}{*}{Publisher} &\multicolumn{2}{c}{NTU RGB+D 120} \\ 
				\cline{3-4} &  & X-sub     &   X-set   \\  
				\midrule
			    ST-LSTM \cite{liu2016spatio}            &ECCV16    &  55.7 & 57.9     \\
				Clips+CNN+MTLN \cite{ke2017new}         &CVPR17    &  62.2 & 61.8     \\
				SkeMotion\cite{caetano2019skelemotion}   &AVSS19    &  67.7 & 66.9     \\ 
				TSRJI  \cite{caetano2019skeleton} &SIBGRAPI19    &  67.9 & 62.8    \\
				\midrule
				ST-GCN \cite{st-gcn}                    &AAAI18    &  70.7 & 73.2     \\
				2s-AGCN \cite{2s-agcn}               &CVPR19    &  82.5 & 84.2     \\
				4s Decouple-GCN \cite{decouple_gcn_cheng2020eccv}   &ECCV2020    &  86.5 & 88.1     \\
                4s Shift-GCN \cite{shift-gcn}   &CVPR2020    &  85.9 & 87.6     \\
				2s MS-G3D \cite{ms-g3d}                     &CVPR20   &  86.9 & 88.4 \\  
                ResGCN \cite{ResGCN}                    &ACMMM20   &  87.3 & 88.3     \\
				4s Dynamic-GCN \cite{dynamic_gcn}                     &ACMMM20   &  87.3 & 88.6 \\ 
				4s MST-GCN \cite{MST-GCN-AAAI2021}                 &AAAI21   & 87.5 & 88.8 \\
				\midrule
				Js DualHead-Net (Ours)         &    -   &  84.6           & 85.9  		\\
				Bs DualHead-Net (Ours)         &    -   &  86.7           & 87.9   		\\
				2s DualHead-Net (Ours)         &    -   &  \textbf{87.9}           & \textbf{89.1}  		\\
				4s DualHead-Net (Ours)         &    -   &  \textbf{88.2}           & \textbf{89.3}  		\\
				\bottomrule
		\end{tabular}}
		\caption{\small Comparison of the Top-1 accuracy (\%)  with the state-of-the-art methods on the NTU RGB+D 120 dataset.}
		\vspace{-0.4cm}
		\label{ntu120}
	\end{table}

		\begin{table}[t!]
		\centering
		\scalebox{1}{
			\begin{tabular}{lcccc}
				\toprule
				\multirow{2}{*}{Methods} & \multirow{2}{*}{Publisher} &\multicolumn{2}{c}{Kinetics-Skeleton} \\ \cline{3-4}
				& & Top-1     & Top-5     \\  \midrule   
				PA-LSTM \cite{ntu_60}               &CVPR16& 16.4           & 35.3            \\ 
				TCN \cite{kim2017interpretable}			&CVPRW17 &20.3	&40.0\\
				ST-GCN \cite{st-gcn}                  &AAAI18& 30.7           & 52.8            \\
				AS-GCN \cite{li2019actional}                  &CVPR19& 34.8           & 56.5            \\
				2s-AGCN \cite{2s-agcn}                  &CVPR19& 36.1           & 58.7            \\
				DGNN \cite{dgcn}              &CVPR19 & 36.9           & 59.6            \\
				NAS-GCN \cite{peng2020learning}          &AAAI20& 37.1           & 60.1            \\
				2s MS-G3D \cite{ms-g3d}                     &CVPR20  &    38.0 & 60.9 \\    
                STIGCN \cite{huang2020spatio}        &ACMMM20   &  37.9 & 60.8     \\
				4s Dynamic-GCN \cite{dynamic_gcn}       &ACMMM20  &    37.9 & \textbf{61.3} \\  
				4s MST-GCN \cite{MST-GCN-AAAI2021}           &AAAI21 & 38.1 & 60.8 \\
				
				\midrule
			    Js DualHead-Net (Ours)         &   -    & 36.6         &  59.5		\\
			    Bs DualHead-Net (Ours)         &   -   &  35.7        &  58.7		\\
			    2s DualHead-Net (Ours)         &   -   &  \textbf{38.3}  & \textbf{61.1} \\
			    4s DualHead-Net (Ours)         &   -   & \textbf{38.4}   & \textbf{61.3}	\\
				\bottomrule
		\end{tabular}}
		\caption{\small Comparison of the Top-1 accuracy (\%) and Top-5 accuracy (\%) with the state-of-the-art methods on the Kinetics Skeleton dataset.}
		\vspace{-0.4cm}
		\label{kinetics}
	\end{table}

\subsection{Ablation Studies}


In this subsection, we perform ablation studies to evaluate the effectiveness of our proposed modules and attention mechanism. Except for the experiments in \textbf{Incremental ablation study}, all the following experiments are performed by modifying the target component based on the full model. Unless stated, all the experiments are conducted under X-sub setting of NTU-RGBD 60 dataset, using only joint stream.

\subsubsection{\textbf{Incremental ablation study}}
We first evaluate the proposed dual head module, temporal and spatial attention mechanism in an incremental manner. We start from our baseline network, MS-G3D\cite{ms-g3d}. We add our proposed modules one-by-one. The results are shown in Table \ref{tab:ablation_full}. Our \textbf{dual head structure} improves the performance from 89.4 to 89.9, which demonstrates the effectiveness such divide-and-conquer structure. Note that our dual head structure keeps less parameters than baseline network due to block simplification. Adding attentions further improves the performance.

\begin{table}[t]
    \centering
    \scalebox{1.0}{
    \begin{tabular}{c|ccccc}
    \hline
      Method & Params & Dual head & TA & SA & Acc \\
      \hline
      Baseline & 3.2M &- & - & - & 89.4 \\
      \hline
      \multirow{3}{*}{Ours}& 3.0M &\checkmark & - & - & 89.9 \\
      & 3.0M & \checkmark & \checkmark & - & 90.2 \\
      & 3.0M & \checkmark & \checkmark & \checkmark & \textbf{90.3} \\
      \hline
    \end{tabular}}
    \caption{\small Ablation study of different modules on NTU RGB+D 60 X-sub setting, evaluated with only joint stream. `TA' indicates cross head temporal attention, `SA' indicates cross head spatial attention. Note that both attention block introduces parameters fewer than 0.1M.}
    \label{tab:ablation_full}
\end{table}

\begin{table}[t]
    \centering
    \scalebox{1.0}{
    \begin{tabular}{c|c|c}
    \hline
       Channel reduction  & Params & Acc \\
       \hline
       Baseline (MS-G3D\cite{ms-g3d}) & 3.2M & 89.4 \\
       No reduction & 4.9M & 90.5 \\
       Reduce channels to 1/2 (proposed) & 3.0M & 90.3 \\
       Reduce channels to 1/4 & 2.5M & 89.7 \\
       \hline
    \end{tabular}}
    \caption{\small Different reduction rate of feature channels in fine head.}
    \vspace{-0.2cm}
    \label{tab:channel_reduction}
\end{table}

\begin{table}[t]
    \centering
    \scalebox{1.0}{
    \begin{tabular}{c|c|c}
       \hline
       G3D pathways & Params & Acc \\
       \hline
       w/o G3D(factorized) & 2.4M & 90.0 \\
       1 G3D & 3.0M & 90.3 \\
       2 G3D & 3.9M & 90.0 \\
       \hline
    \end{tabular}}
    \caption{\small Comparison of different number of G3D pathways in coarse block.}
    \label{tab:g3d_ablation}
    \vspace{-0.2cm}
\end{table}

\subsubsection{\textbf{Model simplification}}
Due to the robust modelling ability of dual head structure, we argue that the GCN blocks in both heads can be simplified for balancing the model performance and complexity. The simplification strategies are investigated and discussed below.



\noindent\textbf{Simplification of fine head.} 
We reduce the channel dimensions of feature maps in fine head due to the rich temporal information contained. In Table \ref{tab:channel_reduction}, we can observe that, without channel reduction, the model achieves an accuracy of 90.5, but with 4.9M parameters. By reducing the channels to 1/2, the accuracy only drops to 90.3, while the parameters are significantly reduced to 3.0M, which is smaller than the baseline network(3.2M). However, as we further reduce the parameters to 2.5M, the accuracy will drop to 87.4. To balance the model complexity and performance, we choose a reduction rate of 2(reduce to 1/2) in our final model.



\begin{figure}[!t]
\includegraphics[width=0.5\textwidth]{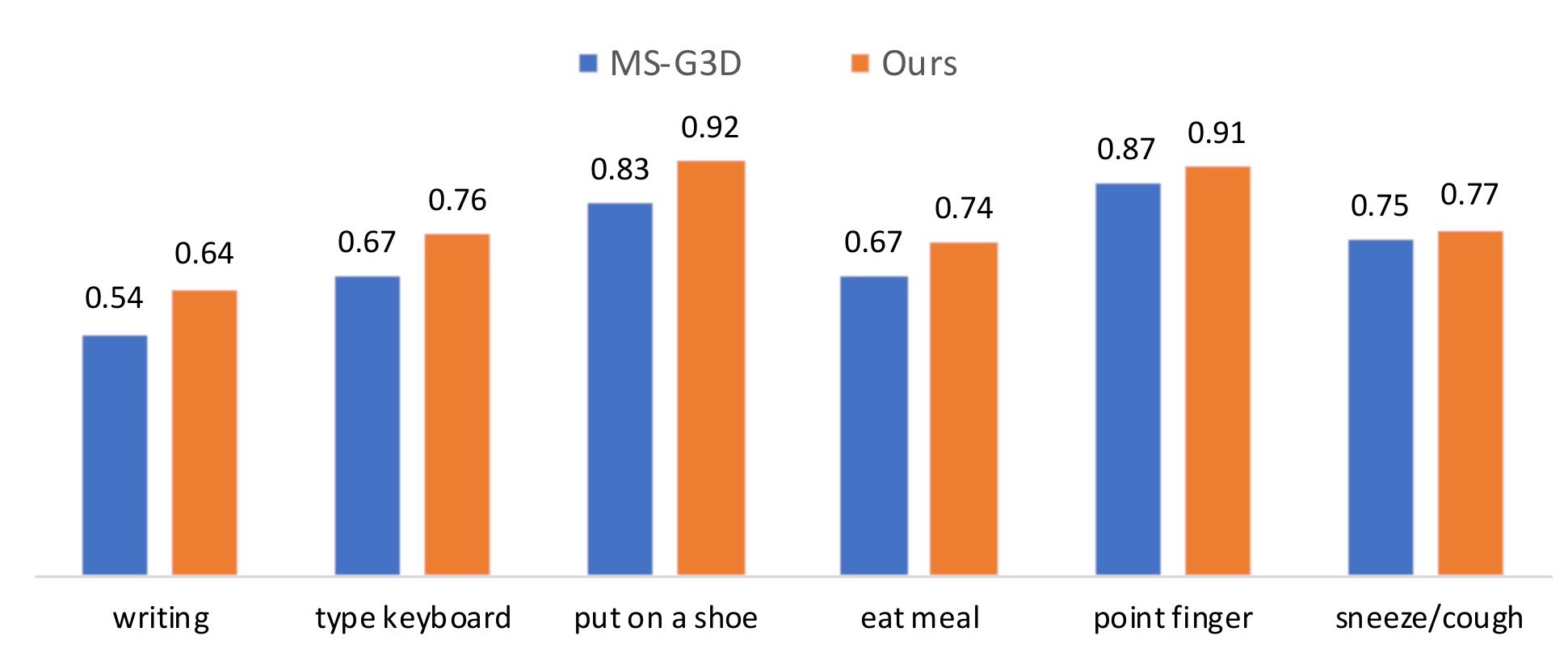}
\vspace{-0.2cm}
\caption{\small Comparison of classification accuracy of 6 difficult action classes on NTU RGB+D X-Sub.}
\label{fig:Comparison_with_g3d}
\Description{}
\end{figure}

\begin{figure}[!t]
	\includegraphics[width=0.45\textwidth]{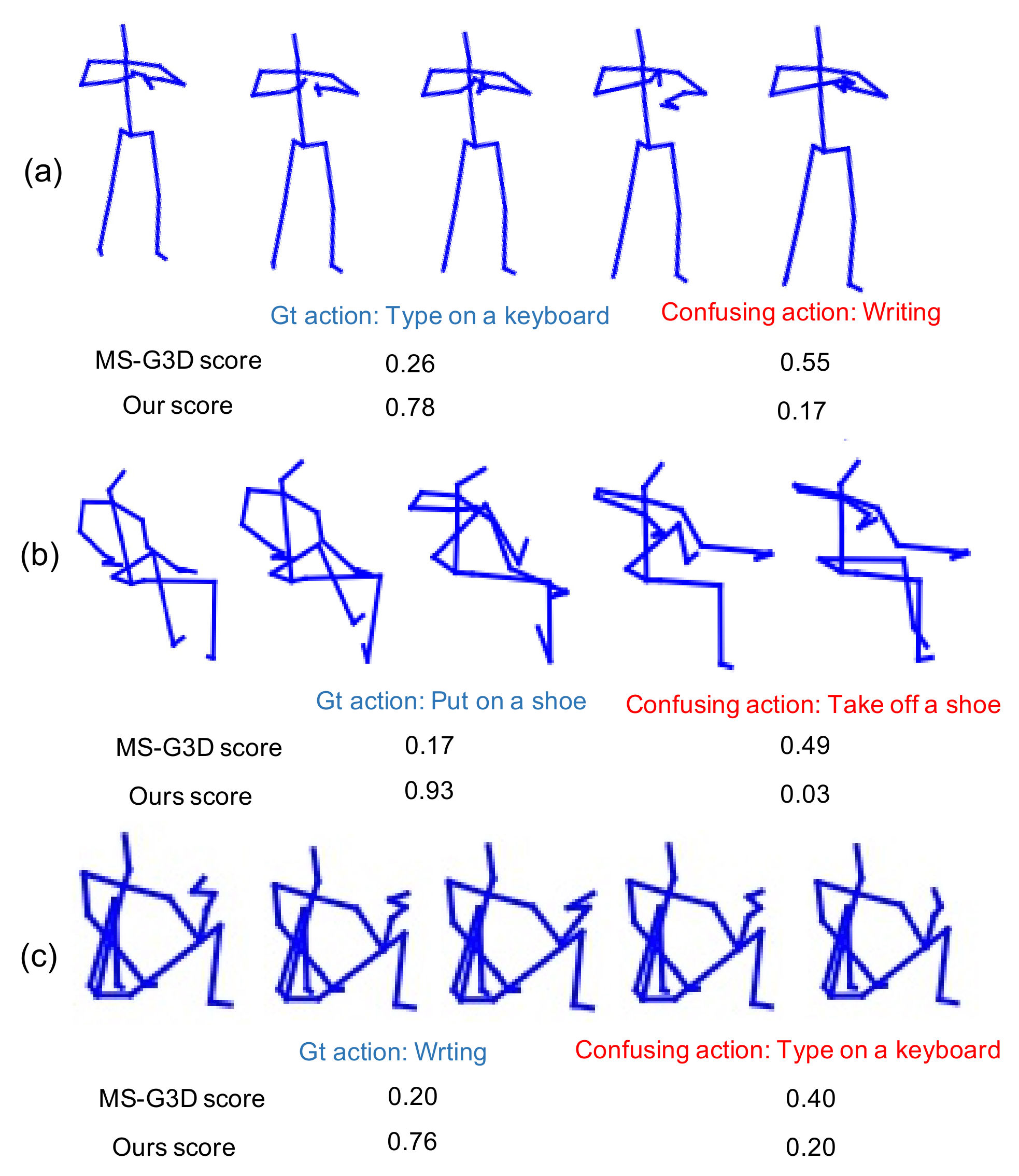}
	\vspace{-0.3cm}
	\caption{\small Skeleton samples and the prediction scores of MS-G3D and our method. GT action and confusing action are shown in blue and red color. Our method improves the prediction scores of those fine-grained actions.}
	\label{fig:Comparison_with_g3d_skeletno}
	\Description{}
\end{figure}


\noindent\textbf{Simplification of coarse head.} We report the ablation study of different G3D pathways in our coarse block in Table \ref{tab:g3d_ablation}. We can observe that utilising one G3D component is able to sufficiently capture the coarse grained motion contexts. Increasing G3D components to 2 will drop the performance a little, we believe it's because the coarse grained motion contexts are easy to capture and large models will turn to over-fitting.

\subsubsection{\textbf{Temporal subsample rate of coarse head.}} We model the coarse grained temporal information in our coarse head, which is generated by subsampling the features in temporal dimension. We hence perform an ablation study of temporal subsampling rate of coarse head, shown in Table \ref{tab:subsampling_rate}. Since proposed method utilise a subsampling rate of 2(reduce to 1/2). We can observe that, without subsampling, the coarse head also takes fine grained features, and the performance will drop from 90.3 to 89.8, demonstrating the importance of coarse grained temporal context. However, as we further enlarge the subsample rate to 4(reduce to 1/4), the performance will drop to 90.1. This implies that over subsampling will lose the important frames and hence drop the performance.

\subsubsection{\textbf{Cross head attention}}
We also perform ablation studies on our proposed cross head temporal attention and spatial attention. Our proposed cross head temporal attention passes fine grained temporal context from fine head to coarse head and re-weight the coarse features. As shown in Table \ref{tab:cross_head_attenion}, such cross head temporal attention mechanism outperforms the attention estimated by coarse features themselves. Similar in Table \ref{tab:cross_head_attenion}, the cross head spatial attention outperforms the spatial attention estimated by fine features, denoted by 'self learned attention'.

\begin{table}[t]
    \centering
    \scalebox{1.0}{
    \begin{tabular}{c|c}
    \hline
       Temporal subsample & Acc \\
       \hline
       subsample all frames & 89.8 \\
       subsample 1/2 frames & 90.3 \\
       subsample 1/4 frames & 90.1 \\
      \hline
    \end{tabular}}
    \caption{Ablation study of temporal subsample rate in coarse head.}
    \label{tab:subsampling_rate}
    \vspace{-0.2cm}
\end{table}

\begin{table}[t]
    \centering
    \scalebox{0.9}{
    \begin{tabular}{c|c|c}
       \hline
       Attention type & Mechanism & Acc \\
       \hline
       \multirow{2}{*}{Temporal attention} & cross head attention& 90.3 \\
       & self learned attention & 90.0 \\
       \hline
       \multirow{2}{*}{Spatial attention} & cross head attention& 90.3 \\
       & self learned attention & 90.1 \\
    \hline
    \end{tabular}}
    \caption{Comparison of cross head attention and self learned attention, which is estimated by the features of their own heads.}
    \label{tab:cross_head_attenion}
    \vspace{-0.2cm}
\end{table}

\subsection{Qualitative Results}
We show some qualitative results in Figure.\ref{fig:Comparison_with_g3d} and Figure.\ref{fig:Comparison_with_g3d_skeletno}. We can observe that our method improves those action classes that are in fine-grained label space, which requires both coarse grained and fine grained motion information to be recognised.


\section{Conclusion}

In this paper, we propose a novel multi-granular spatio-temporal graph network for skeleton-based action recognition, which aims to jointly capture coarse- and fine-grained motion patterns in an efficient way. To achieve this, we design a dual-head graph network structure to extract features at two spatio-temporal resolutions with two interleaved branches. We introduce a compact architecture for the coarse head and fine head to effectively capture spatio-temporal patterns in different granularities. Furthermore, we propose a cross attention mechanism to facilitate multi-granular information communication in two heads. As a result, our network is able to achieve new state-of-the-art on three public benchmarks,namely NTU RGB+D 60, NTU RGB+D 120 and Kinetics-Skeleton, demonstrating the effectiveness of our proposed dual-head graph network.


\begin{acks}
This research is funded through the EPSRC Centre for Doctoral Training in Digital Civics (EP/L016176/1). This research is also supported by Shanghai Science and Technology Program (21010502700).
\end{acks}




\newpage
\balance
\bibliographystyle{ACM-Reference-Format}
\bibliography{sample-base}

\appendix
\end{document}